%% file: paper.tex
\documentclass[]{fairmeta}
\usepackage{silence}
\usepackage[utf8]{inputenc} % allow utf-8 input
\usepackage[T1]{fontenc}    % use 8-bit T1 fonts
\usepackage{url}            % simple URL typesetting
\usepackage{booktabs}       % professional-quality tables
\usepackage{amsfonts}       % blackboard math symbols
\usepackage{nicefrac}       % compact symbols for 1/2, etc.
\usepackage{microtype}      % microtypography
\usepackage{xcolor}
\usepackage{soul}
\definecolor{bluelink}{RGB}{0,113,188}
\definecolor{greenlink}{RGB}{0,188,113}
\definecolor{PineGreen}{RGB}{0.0, 0.47, 0.44}
\definecolor{Gray}{RGB}{0.5,0.5,0.5}
\definecolor{audio_desc}{rgb}{0.82,0.99,0.80}
\definecolor{shot_desc_1}{rgb}{0.98,0.87,0.87}
\definecolor{shot_desc_2}{rgb}{0.98,0.95,0.83}
\definecolor{shot_desc_3}{rgb}{0.92,0.86,0.98}
\usepackage{listings} 
\usepackage{wrapfig}
\usepackage[most]{tcolorbox}

\newtcolorbox{samplebox}[1]{
    breakable, 
    colback=blue!5!white,
    colframe=blue!50!black, 
    fonttitle=\bfseries,
    title=#1
}
\definecolor{citecolor}{HTML}{0071bc}
\hypersetup{
    colorlinks=true,%
    citecolor=citecolor,%
    filecolor=citecolor,%
    linkcolor=citecolor,%
    urlcolor=citecolor
}
\usepackage{tabularx}
\usepackage[most]{tcolorbox}
\usepackage{amsmath}
\usepackage{multirow}
\usepackage{array}
\usepackage{wrapfig}
\usepackage{enumitem}
\usepackage{tikz}
\usepackage{lipsum}
\usepackage{adjustbox}
\usepackage[table]{xcolor}  
\usepackage{siunitx}          
\usepackage{graphicx}     
\usepackage{float}

\usepackage{amssymb}
\renewcommand{\paragraph}[1]{\vspace{1.25mm}\noindent\textbf{#1}}

\newlength\savewidth
\newcolumntype{x}[1]{>{\centering\arraybackslash}p{#1pt}}
\newcolumntype{y}[1]{>{\raggedright\arraybackslash}p{#1pt}}
\newcolumntype{z}[1]{>{\raggedleft\arraybackslash}p{#1pt}}

\usepackage{pgf}
\usepackage{colortbl}
\usepackage{tipa}
\usepackage{tcolorbox}
\usepackage{rotating}
\usepackage[abs]{overpic}
\usepackage{makecell}
\usepackage{longtable}
\usepackage[english]{babel}
\usepackage{csquotes}
\usepackage{hyperref}
\usepackage[most]{tcolorbox}
\definecolor{eventcolor}{RGB}{70,130,180}   % 蓝色（steelblue）
\definecolor{shotcolor}{RGB}{218,165,32}    % 金黄色（goldenrod）
\definecolor{entitycolor}{RGB}{220,20,60}   % 红色（crimson）
\definecolor {globalcoloar}{RGB}{150,210,120}   % 绿色（crimson）

\title{Script-a-Video: Deep Structured Audio-visual Captions via Factorized Streams and Relational Grounding}

\author[]{Tencent Hunyuan Team}

\newcommand{\papertitle}{Script-a-Video: Deep Structured Audio-visual Captions via Factorized Streams and Relational Grounding}

\usepackage{fancyhdr} 
\fancypagestyle{titlepagewithlogo}{
  \fancyhf{} % 清空所有当前的页眉页脚设置
  \rhead{\includegraphics[height=0.6cm]{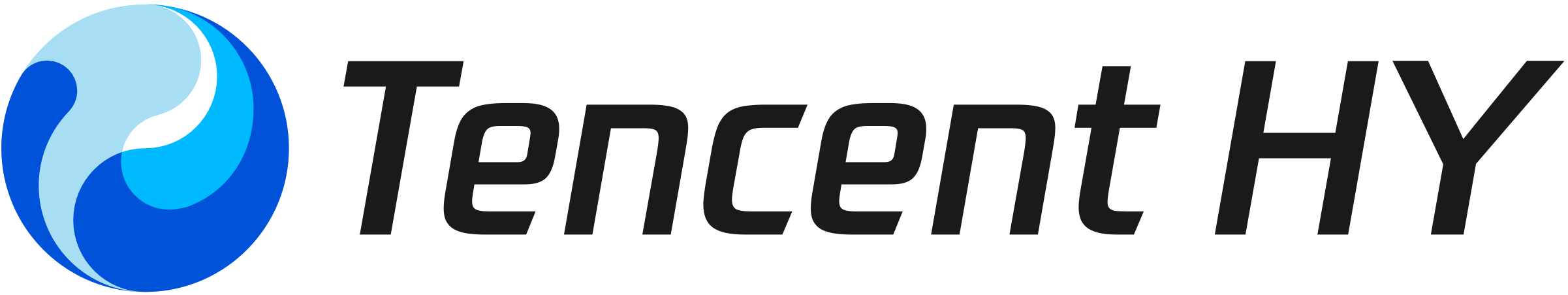}} 
}

\abstract{} % 解决 fairmeta 模板编译报错的问题

\begin{document}
\thispagestyle{titlepagewithlogo}
\maketitle
% \mymaketitle
\vspace{-15mm}
\begin{center}
  \includegraphics[width=\linewidth]{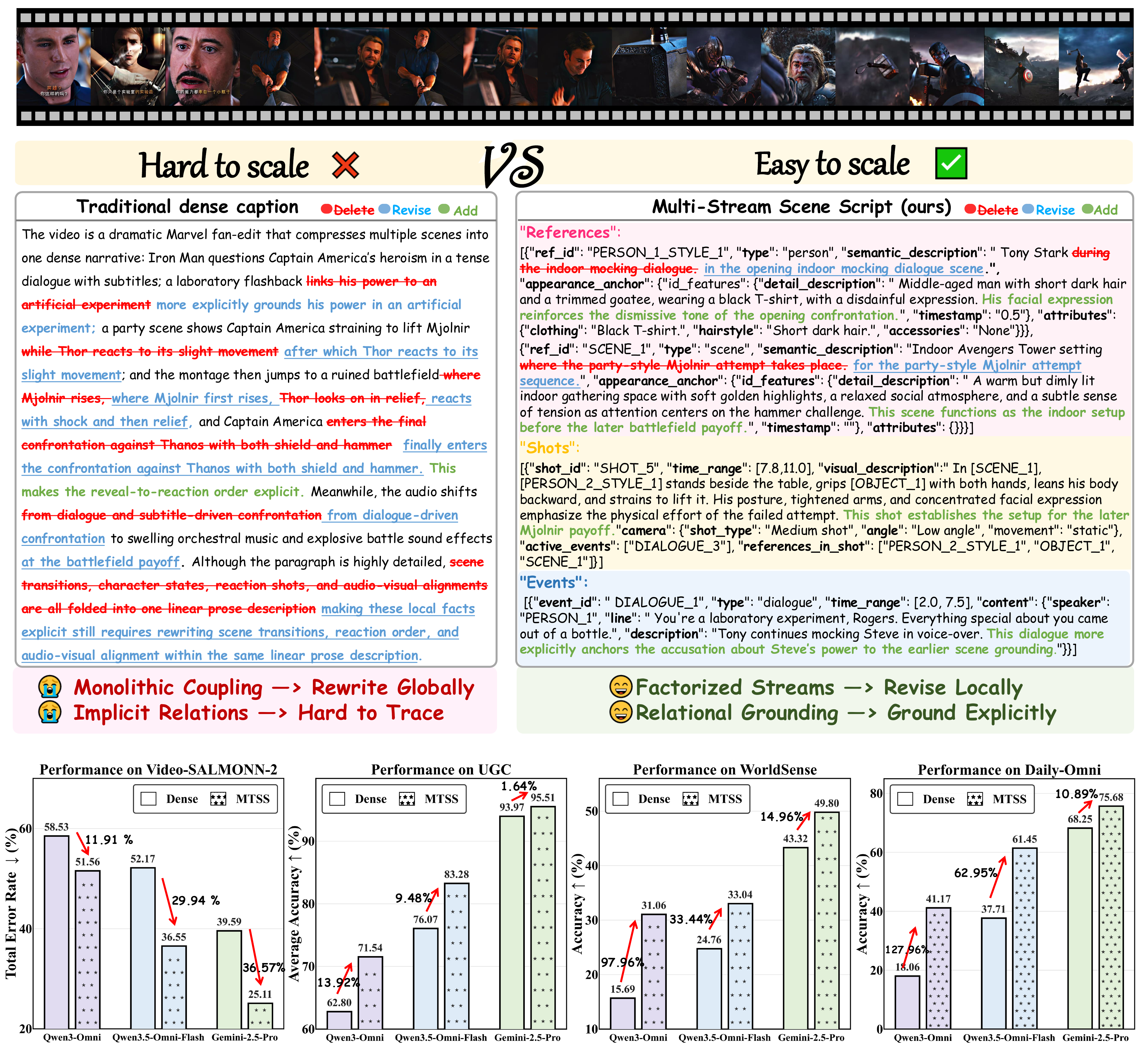}
  \captionsetup{hypcap=false}
  \captionof{figure}{
    \textbf{From monolithic captions to grounded scripts.}
    \textbf{Top (Scalability)}: Monolithic captions entangle visual and auditory elements into a single paragraph, where local edits inevitably trigger global rewrites. MTSS instead factorizes the video into specialized streams connected by explicit relational anchors, making dependencies traceable and enabling precise local updates.
    \textbf{Bottom (Fidelity \& Learnability)}: This factorized structure yields consistent downstream improvements.   
    \textbf{ (i) Fidelity}: By preserving cross-modal relations more faithfully, MTSS lowers error rates on Video-SALMONN-2 and delivers a 110\% gain on the Daily-Omni reasoning task. \textbf{(ii) Learnability}: MTSS also narrows the performance gap between smaller and larger MLLMs, indicating a substantially more learnable caption interface.
}
  \label{fig:teaser_mtss}
\end{center}

\newpage
\begin{center}
{\large\textbf{Abstract}}
\end{center}
Advances in Multimodal Large Language Models (MLLMs) are transforming video captioning from a descriptive endpoint into a semantic interface for both video understanding and generation. However, the dominant paradigm still casts videos as monolithic narrative paragraphs that entangle visual, auditory, and identity information. This dense coupling not only compromises representational fidelity but also limits scalability, since even local edits can trigger global rewrites. To address this structural bottleneck, we propose Multi-Stream Scene Script (MTSS), a novel paradigm that replaces monolithic text with factorized and explicitly grounded scene descriptions. MTSS is built on two core principles: Stream Factorization, which decouples a video into complementary streams (Reference, Shot, Event, and Global), and Relational Grounding, which reconnects these isolated streams through explicit identity and temporal links to maintain holistic video consistency. 
Extensive experiments demonstrate that MTSS consistently enhances video understanding across various models, achieving an average reduction of 25\% in the total error rate on Video-SALMONN-2 and an average performance gain of 67\% on the Daily-Omni reasoning benchmark. It also narrows the performance gap between smaller and larger MLLMs, indicating a substantially more learnable caption interface. Finally, even without architectural adaptation, replacing monolithic prompts with MTSS in multi-shot video generation yields substantial human-rated improvements: a 45\% boost in cross-shot identity consistency, a 56\% boost in audio-visual alignment, and a 71\% boost in temporal controllability.

\input{1_intro_v2}

\input{2_related}

\input{3_method}

\input{4_exp}

\input{5_conclusion}

\clearpage
\appendix
\section*{Project Contributors}
\begin{itemize}[leftmargin=*]
    \item \textbf{Project Sponsors:} Linus
    
    \item \textbf{Project Leader:} Qinglin Lu, Biao Wang
    
    \item \textbf{Core Contributors (Listed alphabetically):} Hongmei Wang, Jiale Tao, Ruitao Chen, Shiyuan Yang, Wenqing Yu, Yuan Zhou, Zhangquan Chen, Zixiang Zhou

    \item \textbf{Contributors (Listed alphabetically):} Bing Wu, Jinzheng Zhao, Linqing Wang, Pengcheng Guo, Rui Chen, Ruihuang Li, Shan Yang, Shisheng Huang, Shuai Shao, Shunkai Li, Weiting Guo, Yifeng Ma, Yufeng Zhang, Zhongyu Yang

\end{itemize}

% \input{appendix}
% \clearpage
% \newpage

\label{references}
\addcontentsline{toc}{section}{References}
\bibliographystyle{assets/plainnat}
\bibliography{paper}

\end{document}

%% file: 1_intro_v2.tex
\section{Introduction}
\label{section:intro}

Recent progress in Multimodal Large Language Models (MLLMs) has markedly advanced both video understanding and generation, enabling models to jointly process visual dynamics, acoustic signals, and linguistic semantics within a unified framework.
As these capabilities mature, video captioning is no longer merely a descriptive endpoint; it increasingly serves as a semantic interface.
For understanding, it distills complex audio-visual scenes into structured evidence for reasoning and retrieval; for generation, it provides executable specifications for identity-consistent and temporally coherent synthesis.
This dual role imposes two demands on caption representations: they must faithfully preserve scene content and cross-modal relations, while remaining modular enough to support efficient refinement and downstream use.

Despite these advances, the dominant captioning paradigm still casts videos as monolithic narrative paragraphs that entangle visual, auditory, and identity information within a single free-form text stream.
Such entanglement inherently weakens the representation before it even reaches the model.
When scene transitions, recurring entities, and concurrent audio-visual events are folded into the same paragraph, the caption becomes locally ambiguous and incomplete. 
Without explicit entity references, recurring characters and scenes must be re-described across shots, introducing redundancy and increasing the risk of inconsistent references and hallucinated identity details.
The same dense coupling also undermines scalability.
Even a local refinement, such as clarifying a camera motion or an acoustic event, often requires rewriting the surrounding context to preserve narrative coherence, making the caption costly to extend and difficult to reuse downstream.
Consequently, these structural flaws not only compromise fidelity and scalability, but also make these representations harder for MLLMs to learn from, especially for smaller models in our empirical analysis.

\begin{figure*}[t]
    \centering
    \includegraphics[width=\linewidth]{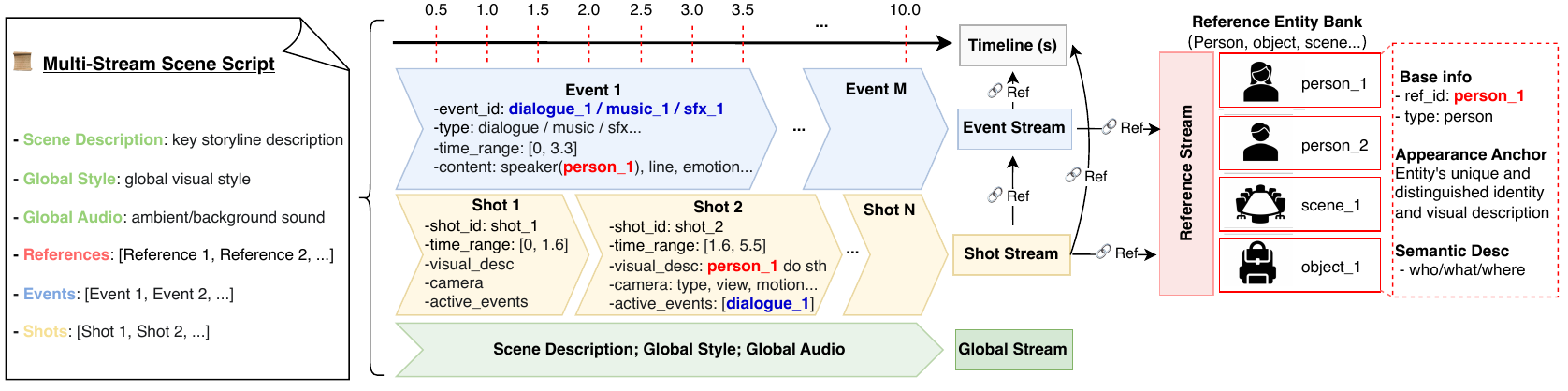} 
    \caption{
    Overview of the Multi-Stream Scene Script (MTSS) representational design.
    \textbf{Stream Factorization} decouples the video into four complementary streams: an \textcolor{eventcolor}{\textbf{Event stream}} for continuous auditory events, a \textcolor{shotcolor}{\textbf{Shot stream}} for discretized visual segments, a \textcolor{entitycolor}{\textbf{Reference stream}} for persistent identities, and a \textcolor{globalcoloar}{\textbf{Global stream}} for shared ambient context.
    \textbf{Relational Grounding} reconnects these isolated streams via explicit identity links (a centralized entity bank) and temporal links (shared anchors and timestamps). 
    Ultimately, MTSS weaves the decoupled streams into a cohesive script, fundamentally improving both fidelity and scalability.
    }
    \label{fig:intro_mtss}
\end{figure*}

We address this bottleneck with \textbf{M}ulti-S\textbf{t}ream \textbf{S}cene \textbf{S}cript (MTSS), a grounded scene-script representation that replaces monolithic captions with structured, referenceable components. 
As illustrated in Figure~\ref{fig:teaser_mtss}, instead of compressing visual and auditory dynamics into a monolithic paragraph where local edits trigger global rewrites, MTSS disentangles these elements so that the representation becomes easier to refine and more faithful in reconstruction.
This dual advantage comes from a different representational design, whose internal structure is detailed in Figure~\ref{fig:intro_mtss}.
MTSS begins with \emph{Stream Factorization}, which organizes a video into four complementary streams: a Reference stream for persistent entities and scenes, a Shot stream for temporally bounded visual segments, an Event stream for localized audio events, and a Global stream for shared ambient context.
By separating persistent information from temporally evolving elements, this factorization reduces semantic redundancy and enables local updates, while preserving the fine-grained cues needed for faithful audio-visual alignment.
Consequently, these descriptions become easier for MLLMs to learn from, especially for smaller models that would otherwise need to disentangle densely interleaved relations from free-form text.
Factorization alone is not sufficient, since isolated streams would fail to form a coherent script.

To bridge these isolated components, MTSS introduces \emph{Relational Grounding}, which reconnects the factorized streams through explicit identity and temporal links. On the identity side, a centralized reference bank assigns persistent identifiers to recurring characters, objects, and scenes, allowing Shot and Event descriptions to refer to the same entities consistently. 
This explicit referencing prevents redundant re-description and keeps identity cues unambiguous across shots. On the temporal side, shared anchors align visual segments with concurrent audio events, while intra-description timestamps capture finer correspondences among actions, utterances, and scene changes. 
Ultimately, these identity and temporal links turn the factorized streams into a cohesive script that preserves fidelity without sacrificing scalability, allowing new details to be added locally without disturbing the global narrative.

Together, these gains in fidelity and scalability translate into consistent improvements in both video understanding and generation.
On captioning benchmarks, MTSS lowers Total Error Rate on Video-SALMONN-2~\citep{videosalmonn2} from 0.5853 to 0.5156 and raises the audiovisual detail score on UGC-VideoCap~\citep{ugcvideocaptioner} from 62.80 to 71.54 relative to Qwen3-Omni~\citep{qwen3omni}, confirming that the factorized script preserves scene content and cross-modal relations more faithfully.
These structural advantages analogously enhance reasoning capabilities: on Qwen3-Omni, MTSS achieves a 127\% performance gain on Daily-Omni~\citep{zhou2025daily} and increases the WorldSense~\citep{hong2025worldsense} score from 0.1569 to 0.3106, indicating that explicit relational grounding helps models recover the event logic needed for downstream inference.
As shown in Figure~\ref{fig:teaser_mtss}, MTSS also narrows the gap between smaller and larger MLLMs, suggesting that a better-structured caption interface makes learning substantially easier across model scales.
Beyond understanding, MTSS also serves as a scalable control interface for personalized, multi-shot, and joint audio-visual generation.
In personalized, multi-shot, and joint audio-visual synthesis, replacing monolithic prompts with MTSS improves cross-shot identity consistency, temporal controllability, and audio-visual synchronization; in multi-shot generation, it raises human-rated consistency from 1.22 to 1.77, audio-visual alignment from 1.18 to 1.85, and multi-shot controllability from 1.00 to 1.71 even before architectural adaptation, showing that structured prompting alone already provides a substantially stronger control interface than monolithic baselines.

In summary, our contributions are as follows:
\begin{itemize}
\item We introduce Multi-Stream Scene Script (MTSS), which, to our knowledge, is the first deeply structured audio-visual scripting paradigm to replace monolithic video captions with factorized and explicitly grounded scene descriptions.
\item We propose two core design principles---\emph{Stream Factorization} and \emph{Relational Grounding}---which disentangle complex audio-visual dynamics into specialized streams and reconnect them through persistent identity and temporal links.
\item We show that MTSS provides a more learnable interface for MLLMs, yielding both more faithful video understanding and more controllable multi-shot audio-visual synthesis.
\end{itemize}

%% file: 2_related.tex
\section{Related Work}
\label{section:related}
\subsection{MLLM for video understanding}
\paragraph{Audio-Visual Understanding.}
The integration of audio and visual modalities better reflects the multisensory nature of real-world perception and requires models to build unified representations over tightly coupled heterogeneous signals.
Accordingly, audio-visual understanding has emerged as an increasingly important direction in MLLMs.

Early video MLLMs were predominantly designed for silent visual understanding~\citep{videollama, videochatgpt, SVAgent}, while later adaptations often treated audio as a simple textual add-on or a weak auxiliary~\citep{li2025videochat, omnivideor1}.
Such superficial fusion fails to capture the intricate temporal and semantic coupling between modalities, thereby limiting holistic scene comprehension. To address these, subsequent works have moved toward more tightly integrated cross-modal architectures and training strategies. MiniCPM-o~\citep{minicpmo} and Baichuan-Omni-1.5~\citep{baichuanomni} augment vision-language foundations with audio capabilities through progressive multi-stage training. Ola~\citep{ola} improves modality alignment by treating video as a central bridging signal across modalities. The Video-SALMONN series~\citep{videosalmonn,videosalmonn2} introduces multi-resolution causal modeling for joint audio-video processing. More recently, Qwen3-Omni~\citep{qwen3omni} adopts a thinker-talker MoE design for stronger audio-visual reasoning, while Uni-MoE-2.0-Omni~\citep{unimoe2} explores dynamic-capacity MoE with progressive fine-tuning and demonstrates strong generalization across a wide range of benchmarks.

Benchmarking efforts have evolved in parallel. JointAVBench~\citep{jointavbench} and OmniVideoBench~\citep{omnivideobench} provide more rigorous evaluations with strict audio-visual correlation, covering diverse audio conditions, reasoning types, etc.. Beyond architecture design and supervised training, reinforcement learning has also become an important route for enhancing cross-modal reasoning. R1-Omni~\citep{r1omni} applies RLVR to omni-modal emotion recognition; EchoInk-R1~\citep{echoinkr1} extends GRPO to open-world reasoning over audio, vision, and text; Omni-R1~\citep{omnir1} targets long-horizon video-audio reasoning with a dual-system design; HumanOmniV2~\citep{humanomniv2} introduces context-aware and logic-based rewards to improve human-like understanding; and OmniVideo-R1~\citep{omnivideor1} further improves audio-visual reasoning through query-intensive grounding and modality-aware fusion.

Despite this rapid progress, existing advances are still centered primarily on closed-set QA-style evaluation. By contrast, open-set dense video captioning remains far less explored and considerably more challenging, as it places higher demands on fine-grained understanding, cross-modal integration, and benchmark verifiability. Importantly, this capability also serves as a critical bridge from audio-visual understanding to generative video modeling.

\paragraph{Audio-Visual Captioning.}
Video captioning extends video understanding from answer selection or short-form responses to open-ended language generation. With the rise of audio-visual models, recent efforts have begun to move beyond visual-only captioning toward joint audio-visual description. On the visual side, GLaVE-Cap~\citep{glavecap} proposes a global-local aligned framework that integrates vision experts, cross-frame prompting, and a dual-stream TrackFusion module to generate more comprehensive descriptions. In the audio-visual setting, Video-SALMONN~2~\citep{videosalmonn2} combines multi-round direct preference optimization with a caption-quality objective that explicitly encourages completeness and factual consistency. DiaDem~\citep{diadem} focuses on faithful dialogue-centric audiovisual captioning through a difficulty-aware two-stage GRPO strategy, with particular emphasis on speaker attribution and utterance fidelity. UGC-VideoCaptioner~\citep{ugcvideocaptioner} addresses the distinctive characteristics of short-form user-generated videos through a staged annotation pipeline covering visual-only, audio-only, and joint audio-visual semantics. Omni-Captioner~\citep{omnicaptioner} further proposes an agentic data engine for producing highly detailed multimodal captions while minimizing hallucination, revealing an inherent trade-off between descriptiveness and factual reliability.

These methods substantially improve caption quality in terms of coverage, detail, and faithfulness. However, \emph{such monolithic descriptions are fundamentally limited for complex multi-shot videos} containing multiple entities, scene transitions, overlapping sound sources, and temporally interleaved events. In these settings, \emph{these monolithic, free-form paragraphs often suffer from semantic redundancy, ambiguous referencing, and weak alignment between visual evolution and concurrent audio dynamics.}

\paragraph{Structured Captioning.}
To overcome the limitations of monolithic captioning, recent work has started to introduce structural priors~\citep{du2025vc4vg,yang2024synchronized} into audiovisual description. ASID~\citep{asid} proposes attribute-structured instruction tuning, combining single-attribute and multi-attribute supervision with an automatic verification pipeline that enforces semantic and temporal consistency. TimeChat-Captioner~\citep{timechatcaptioner} further introduces Omni Dense Captioning with a six-dimensional schema and explicit timestamps, enabling more continuous and script-like narration of multi-scene videos.

While these methods mark important progress toward structured captioning, their notion of structure remains relatively shallow. Attributes and modalities are typically represented as isolated parallel fields rather than relationally grounded components. Consequently, three core limitations persist: (i) the absence of persistent identity references for consistent subject tracking across shots; (ii) the lack of explicit synchronization between visual segments and concurrent audio events; and (iii) the absence of cross-stream mechanisms to weave isolated entities, actions, and sounds into a cohesive script. 

The proposed MTSS paradigm addresses these limitations through \emph{deep structuralization}. Instead of merely appending structural tags to text, MTSS \emph{ factorizes videos into specialized streams and reconnects them through explicit relational links.}

\subsection{Functional Video Generation}
The field of video generation is rapidly transitioning from single-shot, visual-only generation toward complex capabilities including \textit{identity personalization}, \textit{audio-visual co-generation}, and \textit{multi-shot composition}. As these axes converge, the demand for prompts capable of expressing intricate spatio-temporal and cross-modal relationships has surged. In the following, we review recent progress along these dimensions.

\paragraph{Identity Customized Video Generation.}
ID-Customized video generation aims to synthesize videos that maintain identity consistency while following text prompts. We categorize current research based on the mechanism used to bind identity features to prompt semantics:
(1) Explicit placeholder binding: These methods represent identities via specific placeholders or anchors within the prompt. Early works such as DreamVideo \citep{dreamvideo} and VideoDreamer \citep{videodreamer}  utilize textual inversion or subject-specific tokens to learn visual concepts. More recently, HunyuanCustom \citep{hunyuancustom} embeds identities using specialized \texttt{<image>} tokens within VLLM-driven templates, while CustomVideo \citep{customvideo} and Movie Weaver \citep{movieweaver}  insert learnable word vectors or anchored tokens to bind visual concepts to precise text positions.
(2) Implicit semantic binding: This approach~\citep{conceptmaster,tao2025instantcharacter,videoalchemist} maps identity features to natural language labels to achieve intuitive interaction. ConceptMaster \citep{conceptmaster} employs a decoupled attention module to associate image embeddings with category labels like ``dog" or ``cat". Similarly, Video Alchemist \citep{videoalchemist} performs subject-level fusion between specific words and image features, while VideoMage \citep{videomage} uses LoRA modules to implicitly link identity features to corresponding entities in the text prompt. Other works like SkyReels-A2 \citep{skyreelsa2}, PolyVivid \citep{polyvivid}, and Phantom \citep{phantom}  leverage cross-modal alignment or native attention mechanisms to fuse identities without special markers.
(3) Spatial routing binding: These methods utilize spatial priors or regional constraints to bind IDs to specific locations. DisenStudio \citep{disenstudio} introduces spatial-decoupled cross-attention to constrain subject features within predefined areas. Ingredients \citep{ingredients} utilizes an ID router to dynamically assign embeddings to spatio-temporal locations. Furthermore, VideoAnydoor \citep{videoanydoor} uses bounding boxes to guide precise object insertion , and ConsisID \citep{consisid} employs frequency decomposition to inject global and local identity features into specific transformer layers.

\paragraph{Joint Audio-video generation.}
Early exploration in joint audio-video generation often utilizes simple captions, as seen in AV-DIT \citep{avdit}, which supports only basic ambient descriptions and lacks the capacity for fine-grained speech control. To address modality interference, works like BridgeDiT \citep{bridgedit} and 3MDIT \citep{3mdit} propose modality-disentangled prompts that separate visual and acoustic descriptions into independent streams; however, their audio branch still does not support human speech synthesis. UniVerse-1 \citep{universe1} complements this by isolating speech content through distinct audio-specific prompts. Further structural refinement is introduced through tagged prompts, exemplified by Ovi \citep{ovi}, which employs specialized tokens to explicitly define dialogue and sound effect boundaries within a single linguistic context. More recent works employ visual-audio interleaved dense captions; for example, LTX-2 \citep{ltx2} and APOLLO \citep{apollo} utilize prompt rephrasing or multi-expert captioning to integrate comprehensive details—ranging from camera language and verbatim transcripts to nuanced acoustic textures—into a unified, high-density instructional format.

\paragraph{Multi-shot Generation.}
Recent work on multi-shot video generation has evolved through several stages. Early approaches often rely on multi-stage planning-based pipelines, where the global story structure or visual layout is first organized and then rendered into coherent visual content, as exemplified by StoryDiffusion \citep{zhou2024storydiffusion} and Captain Cinema \citep{xiao2025captain}. A subsequent line of work formulates the task in a shot-by-shot manner, generating each new shot conditioned on preceding visual context, as in Cut2Next \citep{he2025cut2next}. More recent methods move toward end-to-end multi-shot generation by extending diffusion or transformer architectures to jointly model multiple shots within a unified framework. Some of these methods introduce explicit shot- or segment-aware designs, such as ShotAdapter \citep{kara2025shotadapter}, HoloCine \citep{meng2025holocine}, and Mask$^2$DiT \citep{qi2025mask}, while others emphasize stronger long-context modeling to improve cross-shot coherence, such as Long Context Tuning \citep{guo2025long} and Mixture of Contexts \citep{cai2025mixture}. In parallel, several works target more specialized cinematic settings, including over-the-shoulder dialogue generation in ShoulderShot \citep{zhang2025shouldershot}, transition-aware synthesis in CineTrans \citep{wu2025cinetrans}, and character-centric movie-style generation in MoCha \citep{wei2025mocha}.

Overall, despite differences in task focus, most existing works still rely on \emph{monolithic prompts} or \emph{weakly structured formulations}. Such designs are often sufficient for isolated settings where the objective is controlling a single capability, such as identity preservation, audio generation, or shot-level continuation. However, they become inadequate in complex video generation scenarios involving \emph{multiple subjects}, \emph{multiple shots}, and \emph{dialogue-intensive interactions}. In particular, existing designs lack explicit mechanisms to model \emph{identity continuity across shots}, making it difficult to maintain consistent subject appearances over time. They also struggle to reliably represent \emph{audio-visual temporal synchronization}, especially when speech, actions, and cinematic transitions must be coordinated at fine temporal granularity. 
In contrast, MTSS fundamentally breaks from this paradigm. By deeply structuralizing entities, events, and their temporal interactions into explicit relational streams, MTSS serves as a highly scalable control interface that inherently resolves the ambiguities of complex, multi-shot audio-visual generation.

%% file: 3_method.tex
\section{Deep Structured Audio-visual Scripts}
\label{section:method}
\definecolor{eventcolor}{RGB}{70,130,180}   % 蓝色（steelblue）
\definecolor{shotcolor}{RGB}{218,165,32}    % 金黄色（goldenrod）
\definecolor{entitycolor}{RGB}{220,20,60}   % 红色（crimson）
\definecolor {globalcoloar}{RGB}{150,210,120}   % 绿色（crimson）

\begin{figure*}[t!]
    \centering
    \includegraphics[width=1.0\linewidth]{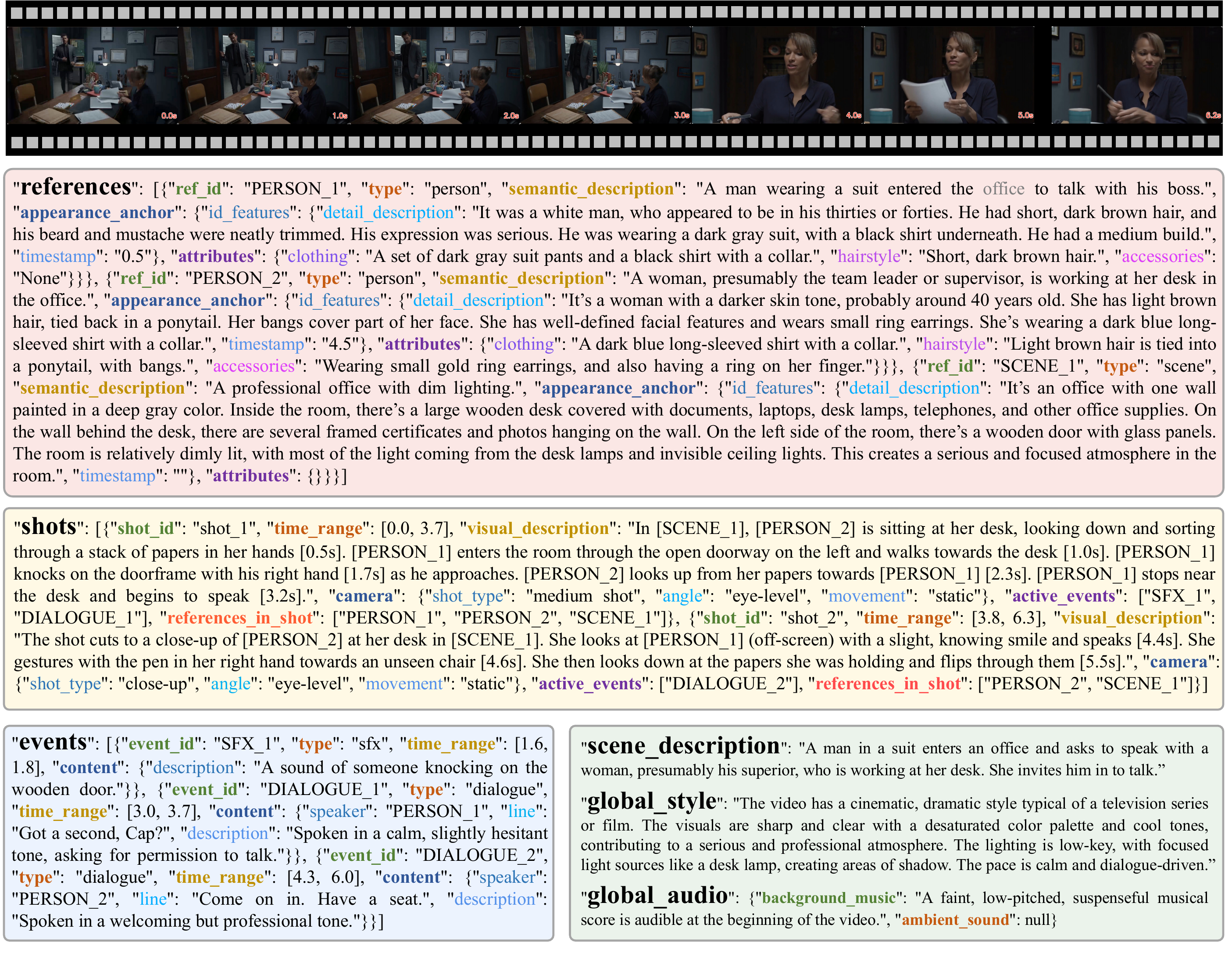} 
    \caption{
    Example of a Multi-Stream Scene Script (MTSS).
    A video is transformed into a structured script composed of four interconnected streams: the \textcolor{entitycolor}{\textbf{Reference stream}} defines a persistent entity bank with identity and appearance anchors; the \textcolor{shotcolor}{\textbf{Shot stream}} organizes visual content into temporally ordered segments with fine-grained descriptions and intra-shot timestamps; the \textcolor{eventcolor}{\textbf{Event stream}} captures temporally grounded auditory events (e.g., dialogue, sound effects) with explicit speaker bindings; and the \textcolor{globalcoloar}{\textbf{Global stream}} provides overall scene context, style, and ambient audio. These factorized streams are unified through shared temporal alignment and reference-based grounding, where shots are linked to their active events and all modalities are grounded to consistent entity identifiers. This structured representation enables non-redundant, identity-consistent, and fine-grained audio-visual modeling across the entire video.
    }
    \label{fig:method_instance}
\end{figure*}

Our framework transforms high-dimensional and entangled video data into a structured, executable script through the synergy of Stream Factorization and Relational Grounding. By factorizing the video into four distinct yet interconnected streams—Reference, Shot, Event, and Global—we create a data representation that is both information-efficient and logically consistent. The essence of MTSS paradigm lies in its ability to decouple complex multimodal information into independent streams while simultaneously re-establishing their semantic and temporal ties, effectively serving as a high-fidelity blueprint for downstream understanding and generation tasks. An example of the script is illustrated in Figure~\ref{fig:method_instance}.

\textbf{The Reference Stream} serves as a foundational Entity Bank that establishes the persistent identities of primary subjects, providing the indispensable WHO for the video narrative and where the narrative unfolds. To maintain narrative focus, this stream selectively factorizes only those entities—categorized as person, object, animal, or scene—that are integral to the main plot, while peripheral elements are relegated to the global scene description. Each entity is defined by a ``semantic\_description'' of its overall state, a ``timestamp'', and an ``appearance\_anchor''. Within the ``appearance\_anchor'', we provide a universal ``detail\_description'' for all entity types; specifically for the person category, we extend this with fine-grained attributes such as clothing, accessories, and hairstyles. This hierarchical design allows subsequent streams to cite persistent Reference IDs instead of repeating descriptions, thereby ensuring absolute identity consistency across the entire script.
\textbf{The Shot Stream} factorizes the visual presentation into a sequence of cinematic segments, organizing visual-spatial content through a rigorous relational lens. Each shot is anchored by a precise ``time\_range'' and encompasses two layers of description. The visual-spatial layer, mirroring the core objectives of traditional video captioning, includes a ``visual\_description'' that provides an objective, chronological narrative of core actions,and a camera field that specifies professional cinematic language (e.g., movements, perspectives, and scales). Crucially, the relational layer binds these descriptions to the broader script: the ``references\_in\_shot'' array maps visible subjects to their Reference IDs, and ``active\_events'' links the shot to concurrent auditory occurrences. Furthermore, we embed intra-description timestamps within the narrative to anchor micro-actions to the global timeline, achieving surgical synchronization between visual content and temporal progression.

\textbf{The Event Stream} achieves audio-visual semantic cohesion by factorizing auditory occurrences that are intrinsically linked to the primary visual narrative. Guided by a strict audio-visual coupling principle, we only extract audio events—classified as dialogue, sfx (sound effect), or music—that have a direct visual counterpart or thematic relevance (e.g., sound effects must be generated by a visible subject). Irrelevant background noise is filtered into the global audio metadata, and simultaneous audio sources are factorized into parallel event entries to ensure clarity. Each event entry contains a ``type'', a ``time\_range'', and a comprehensive content block. This block includes a ``speaker'' (relational binding to a Reference ID), the ``line'' (verbatim text), and a ``description'' capturing nuanced semantics like emotional shifts or vocal techniques. These auditory nuances are precisely aligned with the visual track via micro-level timestamps, ensuring the WHAT of the audio is perfectly grounded on the shared spatio-temporal axis.

\textbf{The Global Stream} provides the macro-level context that encapsulates the overarching atmosphere and environmental setting of the video. While the Reference, Shot, and Event streams focus on fine-grained, factorized details, the Global Stream ensures that the broader semantic landscape is preserved. It includes fields such as ``scene\_description'' to serve as an overall description of the video event, ``global\_style'' to describe the overall aesthetic or genre, and ``global\_audio'' to capture ambient sounds and background music that do not qualify as independent events but contribute to the video's mood. By providing this holistic layer, the Global Stream completes the deep structured script, ensuring that the decoupled detailed streams remain situated within a consistent and well-defined global environment.

%% file: 4_exp.tex
\section{Main Results}
\label{section:exp}

\input{tables/caption_quality}

In this section, we conduct extensive experiments to verify the effectiveness of the proposed MTSS paradigm. Our evaluation is twofold: first, in Section \ref{section:exp.caption_quality}, we assess the Caption Quality to determine if the structured scripts provide a faithful and logically coherent representation of video content. Second, we evaluate the utility for advanced video generation to demonstrate how the scripted priors benefit complex downstream synthesis (detailed in Section \ref{section:exp.results}). 

\subsection{Caption Quality} \label{section:exp.caption_quality}

We evaluate the impact of the MTSS paradigm on two core dimensions: descriptive fidelity and reasoning fidelity. We compare proprietary models (Gemini-2.5-Pro, Qwen3.5-Omni-Flash) and open-source models (Qwen3-Omni, AVoCaDO~\citep{chen2025avocado}, ASID-Captioner~\citep{asid}) using both the native monolithic captions and MTSS-structured scripts. Furthermore, we evaluate the specialized model, Qwen3-Omni-MTSS-FT, which is fine-tuned on the MTSS dataset. To train a model capable of generating MTSS-format captions, we curated an internal dataset comprising 500K high-quality video clips from film, television, and lifestyle domains. Each clip was annotated using Gemini-2.5-Pro. Using this data, we simply performed supervised sine-tuning on the open-source Qwen3-Omni-Instruct~\citep{qwen3omni} model to develop the specialized captioning variant. The quantitative results for audiovisual captioning and downstream reasoning are summarized in Table~\ref{tab:cap_quality} and Table~\ref{tab:cap2qa}, respectively.

\textbf{Evaluation Benchmarks and Metrics.} We evaluate caption quality across two primary tasks: Audiovisual Captioning: We utilize the video-SALMONN-2 test set~\citep{tang2025video-SALMONN_2} to measure error rates, including Missing (Miss), Incorrect, and Hallucinations. Furthermore, we employ UGC-VideoCap~\citep{ugcvideocaptioner} to separately assess the quality of visual and audio tracks, along with a ``Details'' score to measure fine-grained description density. Caption-based Reasoning (VQA): To test the logical consistency and semantic completeness of our scripts, we conduct reasoning tasks on Daily-Omni~\citep{zhou2025daily} and WorldSense~\citep{hong2025worldsense}. In this zero-shot setting, the model must answer complex questions based solely on the generated captions, where higher accuracy indicates a more reconstructible and information-rich script. 

\paragraph{Descriptive Fidelity.}
The experimental results in Table~\ref{tab:cap_quality} demonstrate that MTSS acts as a superior semantic interface that consistently enhances model performance across different scales, even without additional fine-tuning. For proprietary models, switching from monolithic prose to MTSS prompts yields a steady improvement (e.g., Gemini-2.5-Pro's Overall score rises from 93.97 to 95.51 on UGC-VideoCap, and its TotalError improves from 0.3959 to 0.3655 on the video-SALMONN-2 testset).

This enhancement is particularly pronounced for models with smaller parameter scales or lower baseline capabilities. For instance, Qwen3.5-Omni-Flash sees its Overall score jump from 76.07 to 83.28, and its TotalError on Video-SALMONN-2 drops from 0.5217 to 0.3655. This suggests that MTSS-structured scripts significantly lower the cognitive burden of information organization, allowing models to extract and present audio-visual details more effectively by following a structured, form-filling logic rather than juggling complex narrative coherence.

The most significant gains are observed in the fine-tuned Qwen3-Omni-MTSS-FT. As shown in Table~\ref{tab:cap_quality}, it achieves a state-of-the-art Overall score of 85.11 among open-source models, outperforming recent specialized models such as AVoCaDO (83.02) and ASID-Captioner-7B (79.13).
A critical observation lies in the Miss rate on the Video-SALMONN-2 testset: the base Qwen3-Omni exhibits a high omission rate of 0.3734, while the fine-tuned model reduces this to 0.1548. We attribute this leap in comprehensiveness to the fidelity and scalability of the MTSS design. By factorizing the video into decoupled streams, the model learns to populate each field with granular details without the semantic interference common in monolithic paragraph generation. Remarkably, the fine-tuned model narrows the performance gap between open-source and proprietary models, even rivaling the Qwen3.5-Omni-Flash-level proprietary models in descriptive detail.

\paragraph{Reasoning Fidelity.}
The superiority of the grounded scene script is further validated by the audiovisual reasoning results in Table~\ref{tab:cap2qa}. MTSS-based models exhibit a massive performance surge across all reasoning benchmarks. For the open-source Qwen3-Omni, moving to the MTSS format raises its WorldSense score from 0.1569 to 0.3106, and fine-tuning further elevates it to 0.3875—surpassing all existing open-source SOTAs.
We attribute this 97\%+ gain in reasoning accuracy to the relational-grounding mechanism—the explicit identity and temporal ties between streams—which is essential for recovering event logic. Unlike monolithic captions that require the model to re-parse entangled text, MTSS provides a reconstructible blueprint where the WHO, WHERE, and WHEN are already disambiguated. This structural clarity allows downstream reasoning modules to focus on logical inference rather than identity resolution or temporal de-aliasing.

\paragraph{Summary.}
Overall, the results confirm that MTSS provides a more learnable and expressive interface for MLLMs. The consistency of improvements across zero-shot prompting and supervised fine-tuning proves that structured scripting is not merely a data format but a fundamental advancement in how video semantics should be represented to maximize the potential of multimodal models.

% from chris
\subsection{MTSS as a Blueprint for Advanced Video Synthesis}
To explore the downstream potential of the MTSS paradigm, we evaluate its performance across several frontier generative tasks: multi-shot narrative control, identity customization, and joint audio-visual generation. These tasks require sophisticated coordination between structural logic, actor consistency, and temporal synchronization, providing a rigorous testbed for our structured scripts.

We select LTX-2~\citep{hacohen2601ltx} as our base generative framework for two primary technical reasons. First, its Gemma-based VLM encoder is inherently proficient at parsing and interpreting the structured, JSON-like syntax of MTSS, allowing the model to extract fine-grained semantic instructions from factorized streams. Second, its asymmetric dual-stream Diffusion Transformer (DiT) architecture provides a native foundation for joint audio-visual synthesis. This dual-stream design allows us to directly map the synchronized shot and event streams of MTSS into the latent space of both video and audio branches, facilitating a high-fidelity mapping from scripted instructions to multimodal embeddings.

\paragraph{Problem Definition.}
We define the generative task as a conditioned mapping from the structured MTSS triplet $\mathcal{S}_{\text{ref}}, \mathcal{S}_{\text{shot}}, \mathcal{S}_{\text{eve}}$ to a synchronized video-audio pair $\mathcal{V}, \mathcal{A}$. Specifically, $\mathcal{S}_{\text{ref}}$ provides persistent character and scene identities, $\mathcal{S}_{\text{shot}}$ dictates a chronological sequence of cinematic segments and correspoding descriptions, and $\mathcal{S}_{\text{eve}}$ specifies localized auditory events and their corresponding narrative descriptions. The objective is to synthesize a multi-shot sequence where $\mathcal{V}$ and $\mathcal{A}$ exhibit both identity persistence (consistent subjects across shots) and temporal-auditory precision (exact sound-to-action alignment). This unified task requires the model to resolve complex narrative flows and relational constraints within a single, joint generative pass.

\paragraph{Core Challenges.} This task represents a frontier generative scenario requiring the simultaneous satisfaction of three tightly coupled relational constraints: First, \textit{multi-shot temporal and semantic fidelity}: The model must execute precise transitions at $\mathcal{S}_{\text{shot}}$ timestamps while ensuring visual segments faithfully align with their localized narratives. The primary hurdle lies in maintaining global narrative continuity across fragmented temporal segments without context leakage or boundary artifacts \citep{meng2025holocine}. Second, \textit{cross-shot identity persistence}: Mapping the Reference Stream ($\mathcal{S}_{\text{ref}}$) to generated subjects is particularly challenging in multi-reference scenarios where subject confusion is prevalent. Beyond initial identity injection, the model must achieve robust token-level binding to preserve subject traits under dramatic shifts in viewpoint, lighting, and scale \citep{huang2026rethinking, wang2025multishotmaster}. Third, \textit{fine-grained audio-visual synchronization}: Achieving audiovisual realism requires the sub-frame coordination of visual events and their auditory counterparts. This entails synchronizing lip dynamics with the verbatim dialogue in $\mathcal{S}_{\text{eve}}$ while simultaneously anchoring localized sound effects (e.g., footsteps, impacts) to their respective visual actions and scripted time segments \citep{ltx2, ovi, chern2026speed}. Resolving this deep temporal coupling across decoupled modalities is essential for a logically coherent multimodal narrative.

\begin{figure}[ht]
    \centering
    \includegraphics[width=0.95 \textwidth]{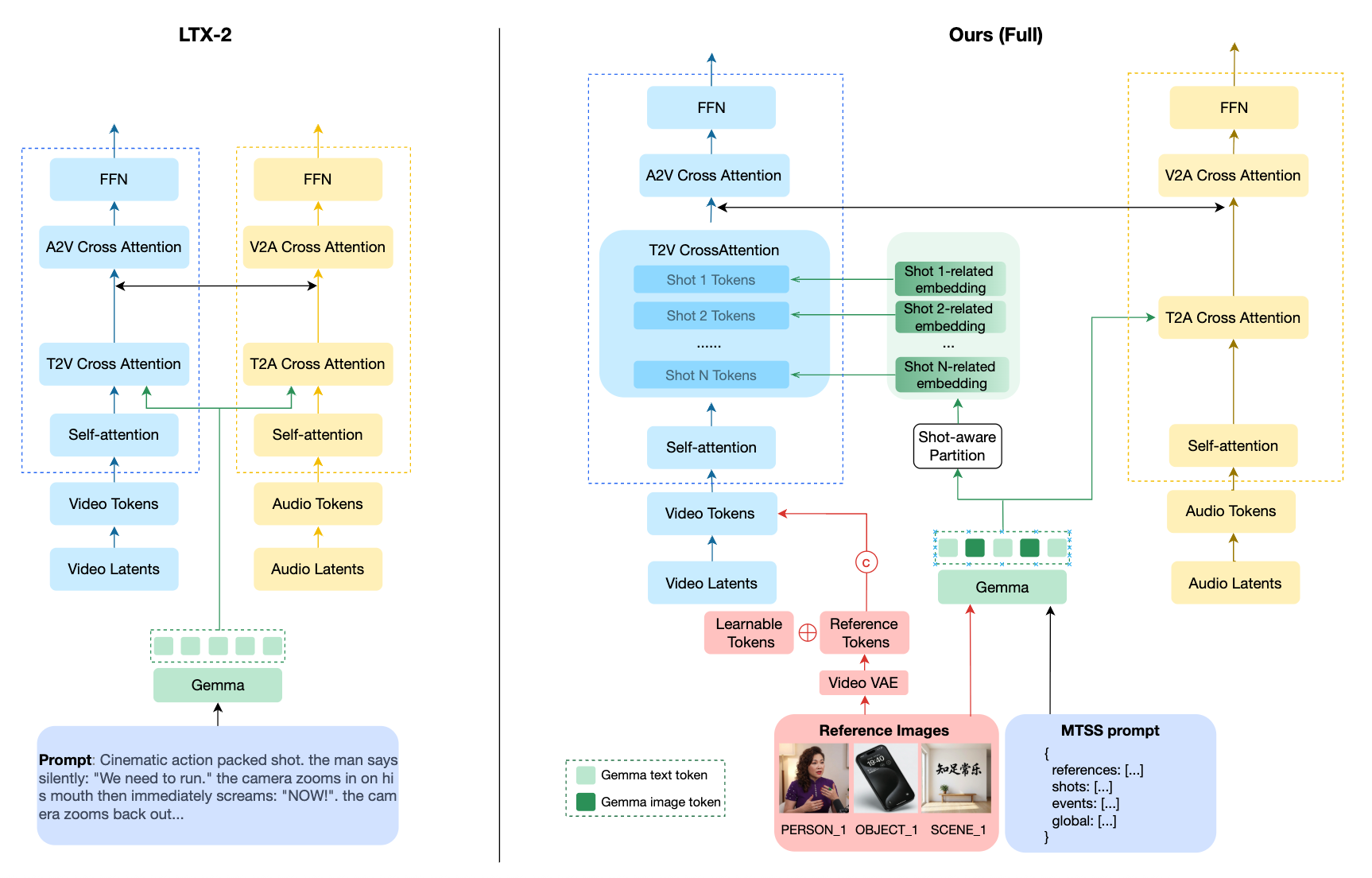}
    \caption{Overview of our full pipeline. Built upon the dual-branch DiT architecture of LTX-2 (left side), our method (right side) introduces two core architectural improvements: (1)~Shot-Aware Structured Attention, which partitions video tokens according to MTSS shot boundaries and performs cross-attention with shot-specific Gemma-3 embeddings for inter-shot context isolation; (2)~Identity Customization, which injects character/object identities via reference VAE features and learnable tokens. Additionally, multi-modal information is fed into Gemma-3 in an interleaved image-text format, providing enriched semantic representations for both the video and audio branches.}
    \label{fig:pipeline}
\end{figure}

\paragraph{Core Architectural Improvements}
We integrated two core improvements on top of the baseline model:
\begin{itemize}
\item \emph{Shot-Aware Structured Attention}: The MTSS Shot Stream inherently provides the precise temporal boundaries and camera language needed for smooth multi-shot transitions. Building upon this, we introduce Shot-Aware Structured Attention. This mechanism first partitions the Gemma-3 embeddings by shot boundaries, then extracts and cross-attends the corresponding video tokens for each shot. By ensuring temporal visual tokens attend only to their respective semantic segments, this explicit temporal factorization effectively prevents inter-shot interference.

\item \emph{Identity Customization}: Traditional video synthesis often struggles with semantic-visual binding ambiguity, where the model fails to strictly associate specific visual features (e.g., from a reference image) with their corresponding textual descriptions in a monolithic prompt. The Reference Stream in MTSS provides a structural solution to this challenge by decoupling entity-specific features into a centralized bank. We leverage this by introducing specialized \emph{reference-learnable-tokens}, which explicitly aligns reference ID tokens (e.g., ``PERSON\_1'') to the corresponding visual image, acting as a relational bridge between the visual identities and their linguistic mentions.
\end{itemize}

For simplicity, we denote Ours (Full) the complete pipeline incorporating all aforementioned improvements, Ours w/o ID removes the identity customization module, Ours w/o MS removes the shot-aware attention partitioning, and Ours w/o AV removes the audio input.

\paragraph{Evaluation Metrics.}
We construct an internal evaluation dataset comprising 125 single-shot samples and 100 multi-shot samples. The data covers a diverse range of categories and scenarios, including movie and TV drama clips, short-form videos, indoor scenes, and outdoor scenes.

Our automated evaluation suite spans three dimensions.
\textit{(i)~Video quality.}
Intra-Shot Subject Consistency (SC) measures temporal appearance coherence within a single shot: we uniformly sample frames, extract DINOv2~\citep{oquab2023dinov2} \texttt{[CLS]} features, and report the mean pairwise cosine similarity.
Reference ID Similarity (Ref.\ ID Sim.) quantifies identity preservation between a reference face image and the generated video: ArcFace~\citep{deng2019arcface} embeddings are extracted from the reference and from sampled video frames, and their mean cosine similarity is reported.
\textit{(ii)~Audio quality.}
Audio Quality is the predicted Mean Opinion Score (MOS) from UTMOS~\citep{saeki2022utmos}, a lightweight speech-quality estimator.
Word Error Rate (WER) evaluates speech content accuracy: we transcribe the generated audio with Whisper-large-v3~\citep{radford2022whisper} and compute the standard WER against the ground-truth transcript using \texttt{jiwer}, with jieba-based tokenization for CJK text.
\textit{(iii)~Audio-visual alignment.}
Semantic Following (Sem.\ Following) evaluates how faithfully the generated video follows the semantic intent of the input prompt, assessed by prompting a Gemma4-31B-it~\citep{gemma_4_2026} vision-language model to independently rate four sub-dimensions---\textbf{Subject} (identity and appearance fidelity), \textbf{Action} (correctness of depicted motions and interactions), \textbf{Scene} (accuracy of background, environment, and spatial layout), and \textbf{Style} (adherence to the requested visual style, color palette, and mood). The overall Semantic Following score is the arithmetic mean of the four sub-scores.
Audio-Video Sync (A-V Sync) quantifies lip-sync accuracy using SyncNet~\citep{chung2016out}: the model outputs a synchronization confidence score and a temporal offset in frames.
Shot Boundary Deviation (Shot Bnd.\ Dev.) measures the absolute frame-level deviation between predicted and ground-truth shot boundaries.

To ensure a rigorous and multi-faceted assessment of the synthesized content, we further conducted a human evaluation involving 20 professional raters. The generated videos were scored across four critical dimensions: Text Alignment (TA), Visual Quality (VQ), Multi-Shot-Consistency (MSC), Identity Consistency (Cons.), and Audio-Video Synchronization (A-V).

% ============================================================
% Section 4.3:  Results and Analysis
% ============================================================

\subsubsection{Results and Analysis}  \label{section:exp.results}

\begin{figure}[t]
    \centering
    \vspace{-0.3cm}
    \includegraphics[width=1.0 \textwidth]{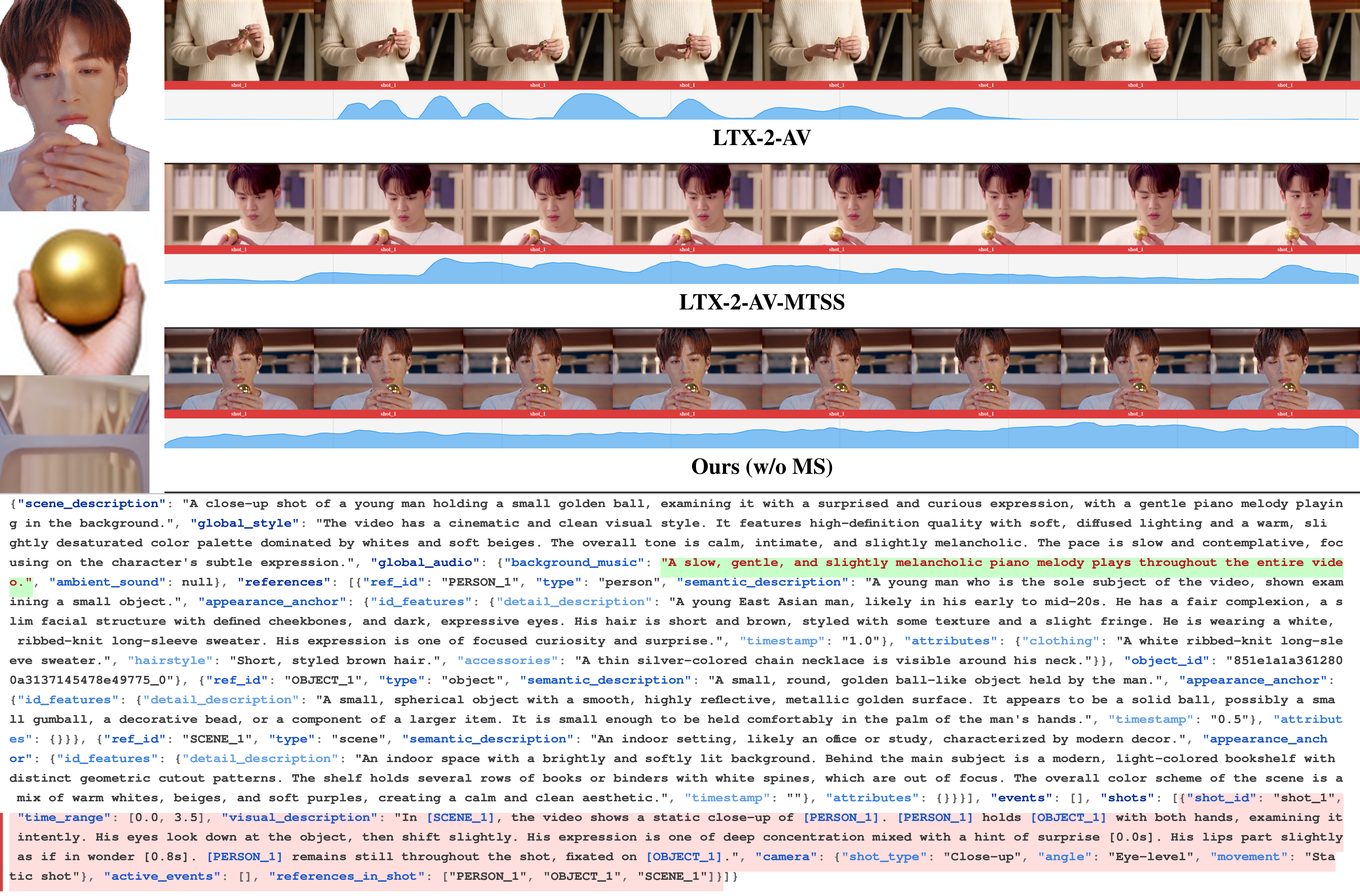}
    \caption{Single-shot qualitative comparison (Example 1). The text specifies a piano melody throughout the video. Our method preserves the reference identity and sustains continuous audio coverage. LTX-2-AV suffers from identity drift and sparse piano fragments; LTX-2-AV-MTSS improves audio RMS coverage over LTX-2-AV, showing that the structured MTSS format alone already benefits audio-visual alignment. \sethlcolor{audio_desc}\hl{Audio description, including background music, ambient sound, and dialog.}\sethlcolor{shot_desc_1}\hl{ Shot description, including time range, visual content description, etc.}}
    \label{fig:single_shot_1}
    \vspace{-0.3cm}
\end{figure}

\begin{figure}[ht]
    \centering
    \includegraphics[width=1.0 \textwidth]{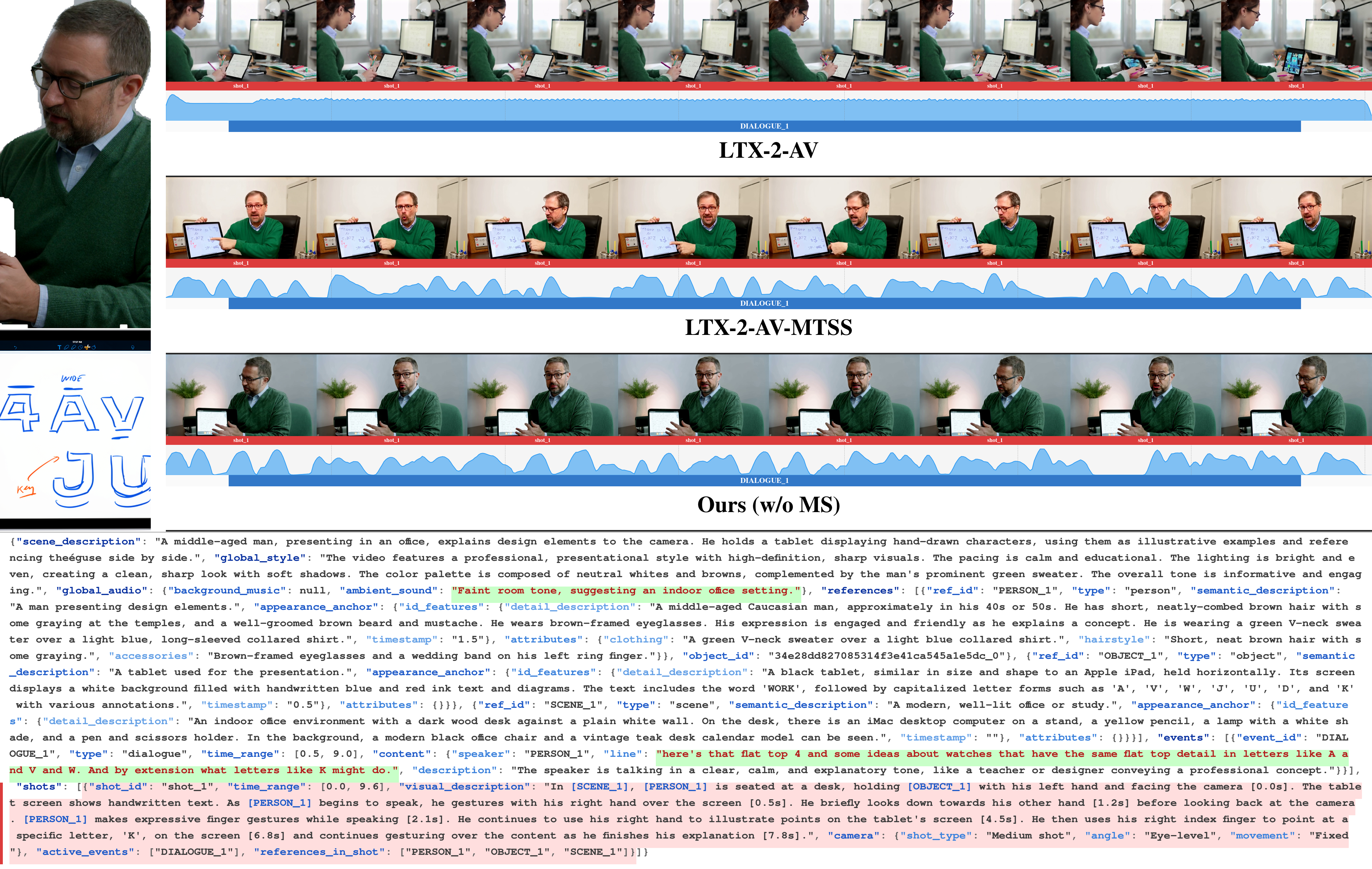}
    \caption{Single-shot qualitative comparison (Example 2). The caption describes a person speaking. LTX-2-AV produces a flat, uniform audio RMS indicative of background noise rather than speech. Both LTX-2-AV-MTSS and our method generate audio RMS with clear speech-like rhythmic patterns, confirming that the MTSS structure guides speech generation. Our full pipeline further achieves precise lip-sync alignment.\sethlcolor{audio_desc}\hl{Audio description, including background music, ambient sound, and dialog.}\sethlcolor{shot_desc_1}\hl{ Shot description, including time range, visual content description, etc.}}
    \label{fig:single_shot_2}
\end{figure}

\begin{figure}[ht]
    \centering
    \includegraphics[width=1.0 \textwidth]{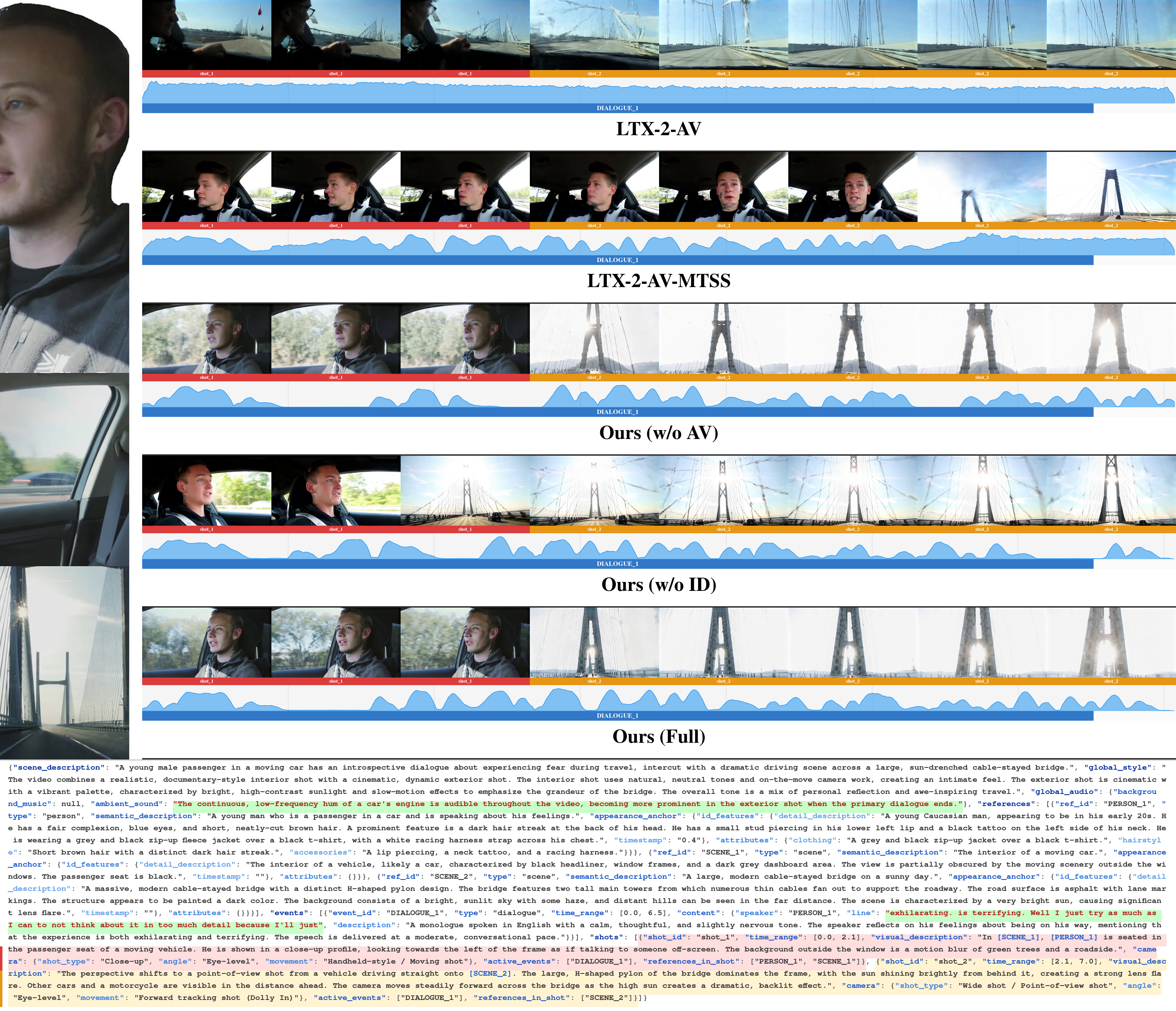}
    \caption{Multi-shot qualitative comparison (Example 1). The caption describes a person speaking across multiple shots. LTX-2-AV fails to reproduce the reference identity and generates flat ambient-noise audio. LTX-2-AV-MTSS improves identity similarity and introduces speech-like audio rhythms, but its shot boundaries deviate from the scripted timestamps. Our variants achieve faithful identity, accurate shot transitions, and speech-aligned audio RMS. \sethlcolor{audio_desc}\hl{Audio description, including background music, ambient sound, and dialog.}\sethlcolor{shot_desc_1}\hl{ Description of shot 1.}\sethlcolor{shot_desc_2}\hl{ Description of shot 2.}}
    \label{fig:multi_shot_1}
\end{figure}

\begin{figure}[ht]
    \centering
    \includegraphics[width=1.0 \textwidth]{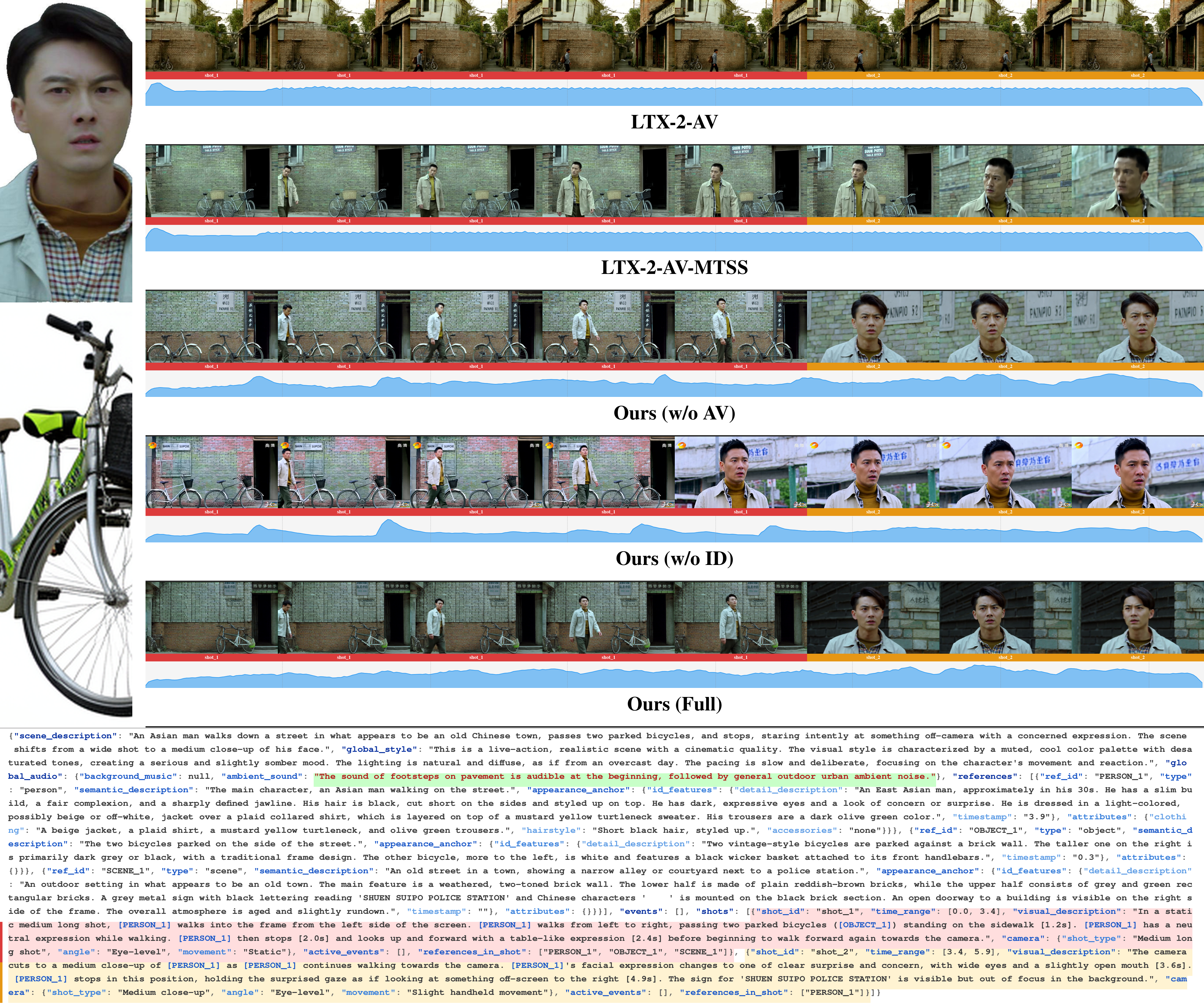}
    \caption{Multi-shot qualitative comparison (Example 2). The caption specifies ``sound of footsteps on pavements'' across shots. LTX-2-AV lacks shot boundaries and exhibits flat audio RMS without footstep patterns. LTX-2-AV-MTSS improves identity consistency but fails to generate shot transitions or synchronized audio. Our variants produce clear shot boundaries, high-fidelity identity, and periodic audio RMS peaks aligned with footsteps. \sethlcolor{audio_desc}\hl{Audio description: music, ambient sound, and dialog.}\sethlcolor{shot_desc_1}\hl{ Description of shot 1.}\sethlcolor{shot_desc_2}\hl{ Description of shot 2.}}
    \label{fig:multi_shot_4}
\end{figure}

We evaluate the MTSS representational paradigm through automated metrics, human evaluation, and qualitative comparisons, comparing against two baselines on the same LTX-2 backbone: (1)~LTX-2-AV, conditioned on monolithic prompts; and (2)~LTX-2-AV-MTSS, conditioned on MTSS scripts \emph{without} architectural enhancements. Since these two baselines share an identical architecture and training paradigm, any performance gap between them is directly and entirely attributable to the MTSS paradigm. We organize our analysis along three axes---Multi-shot Control, Identity Customization, and Audio-Visual Synchronization. Quantitative results are reported in Tables~\ref{tab:auto-metrics} and~\ref{tab:human-eval}; representative qualitative comparisons are shown in Figures~\ref{fig:single_shot_1}--\ref{fig:multi_shot_4}.

% ---------- Automated Metrics Table ----------
\begin{table}[t]
\centering
\small
\caption{Automated quantitative metrics across all configurations. $\uparrow$: higher is better; $\downarrow$: lower is better. Best results per column are in \textbf{bold}. ``--'' indicates that the metric is not applicable to the configuration.}
\label{tab:auto-metrics}
\resizebox{\textwidth}{!}{
\begin{tabular}{lccccccc}
\toprule
\textbf{Method}
  & \textbf{Intra-Shot SC}$\uparrow$
  & \textbf{Ref. ID Sim.}$\uparrow$
  & \textbf{Audio Quality}$\uparrow$
  & \textbf{WER}$\downarrow$
  & \textbf{Sem. Following}$\uparrow$
  & \textbf{A-V Sync}$\downarrow$
  & \textbf{Shot Bnd. Dev.}$\downarrow$ \\
  & {[0,\,1]}
  & {[-1,\,1]}
  & {MOS [1,\,5]}
  & {[0,\,$\infty$)}
  & {[0,\,5]}
  & {[0,\,$\infty$)}
  & {[0,\,$\infty$)} \\
\midrule
\multicolumn{8}{l}{\textit{Single-Shot}} \\
\quad LTX-2-AV                      & --   & --            & 2.83          & 1.64          & 4.12          & 10.04         & --    \\
\quad LTX-2-AV-MTSS                  & --   & --            & 2.50          & 0.78          & 4.79          & 9.69          & --    \\
\quad Ours(w/o MS)              & --   & \textbf{0.62} & 2.46          & 0.23          & 3.88          & 11.68         & --    \\
\midrule
\multicolumn{8}{l}{\textit{Multi-Shot}} \\
\quad LTX-2-AV                      & \textbf{0.87} & --   & 2.32          & 0.84          & 4.18          & \textbf{6.86} & 3.79  \\
\quad LTX-2-AV-MTSS                  & 0.66          & --   & 2.52          & \textbf{0.13} & 4.60 & 13.86         & 3.27  \\
\quad Ours(w/o AV)                   & 0.59          & 0.20 & --            & --            & 4.60            & --            & 1.28  \\
\quad Ours(w/o ID)                     & 0.61          & --   & 2.15          & 0.29          & 4.60          & 9.16          & \textbf{0.38} \\
\quad Ours(Full)              & 0.59          & 0.22 & 2.17          & \textbf{0.19} & \textbf{4.68}          & 9.72          & 1.36  \\
\bottomrule
\end{tabular}
}
\end{table}

% ---------- Merged Human Eval Table ----------
\begin{table}[t]
\centering
\small
\caption{Human evaluation results for single-shot (125 samples) and multi-shot (100 samples) generation. All metrics are on a 1--3 scale ($\uparrow$). Best results per setting are in \textbf{bold}, second best are \underline{underlined}. ``--'' indicates the configuration is not applicable.}
\label{tab:human-eval}
% \resizebox{\textwidth}{!}{
\begin{tabular}{lccccc|cccccc}
\toprule
& \multicolumn{5}{c|}{\textit{Single-Shot}} & \multicolumn{6}{c}{\textit{Multi-Shot}} \\
\cmidrule(lr){2-6} \cmidrule(lr){7-12}
\textbf{Method} & \textbf{TA}$\uparrow$ & \textbf{VQ}$\uparrow$ & \textbf{Cons.}$\uparrow$ & \textbf{A-V}$\uparrow$ & \textbf{Avg.}$\uparrow$ & \textbf{TA}$\uparrow$ & \textbf{VQ}$\uparrow$ & \textbf{Cons.}$\uparrow$ & \textbf{A-V}$\uparrow$ & \textbf{MSC}$\uparrow$ & \textbf{Avg.}$\uparrow$ \\
\midrule
LTX-2-AV         & 1.78 & 2.12 & 1.40 & 1.60 & 1.72 & 1.16 & 1.28 & 1.22 & 1.18 & 1.00 & 1.17 \\
LTX-2-AV-MTSS    & 2.44 & 2.31 & 2.16 & 1.96 & 2.22 & 1.75 & 1.84 & 1.77 & 1.85 & 1.71 & 1.78 \\
\midrule
Ours (w/o MS) & \textbf{2.70} & \textbf{2.50} & \textbf{2.69} & \textbf{2.64} & \textbf{2.63} & -- & -- & -- & -- & -- & -- \\
Ours (w/o ID)           & -- & -- & -- & -- & -- & 2.12 & 2.04 & 2.05 & \underline{2.12} & \textbf{2.62} & 2.19 \\
Ours (w/o AV)     & -- & -- & -- & -- & -- & \underline{2.14} & \underline{2.03} & \underline{2.38} & 2.20 & 2.49 & \underline{2.25} \\
Ours (Full) & -- & -- & -- & -- & -- & \textbf{2.18} & \textbf{2.06} & \textbf{2.41} & \textbf{2.26} & \underline{2.59} & \textbf{2.30} \\
\bottomrule
\end{tabular}
% }
\end{table}

% ============================================================
% Axis 1: Multi-shot Control
% ============================================================

\paragraph{Multi-shot Control.}
As illustrated in Table~\ref{tab:auto-metrics}, LTX-2-AV-MTSS reduces Shot Bnd.\ Dev. from 3.79 (LTX-2-AV) to 3.27 \emph{purely through representational replacement}. Atop this prior, lightweight shot-aware attention further reduces deviation to 0.38 (Ours w/o ID), demonstrating that MTSS shot-boundary timestamps alone provide an effective segmentation signal. The qualitative comparisons in Figures~\ref{fig:multi_shot_1} and~\ref{fig:multi_shot_4} vividly corroborate these metrics. In Figure~\ref{fig:multi_shot_1}, LTX-2-AV-MTSS already shows improvement over LTX-2-AV. The three of our pipeline variants further produce shot transitions that align closely with the specified timestamps, visually confirming the quantitative gains reflected in Shot Bnd.\ Dev. These findings are further validated by human evaluation (as illustrated in Table~\ref{tab:human-eval}), LTX-2-AV scores MSC of only 1.00, LTX-2-AV-MTSS raises it to 1.71 solely through MTSS captions, and our configurations further elevate MSC to 2.49--2.62.

\paragraph{Identity Customization.}
Our evaluation reveals several key insights regarding the synergy between the MTSS Reference Stream and our customization modules.

Firstly, in single-shot scenarios, our framework (Ours w/o MS) achieves a Ref. ID Sim. of 0.62, this high absolute similarity validates that the Reference Stream provides a robust semantic foundation for precise pixel-level identity alignment. While the automated Ref. ID Sim. drops to approximately 0.22 in multi-shot settings—largely due to extreme viewpoint shifts and lighting variations—human evaluations and qualitative evidence (Figure~\ref{fig:multi_shot_1} and Figure~\ref{fig:multi_shot_4}) confirm its effectiveness. Human-rated Cons. scores remain above 2.40, whereas the LTX-2-AV baseline fails entirely to maintain subject identity. Notably, we observe a slight increase in Shot Bnd. Dev. (from 0.38 to 1.36) when identity injection is enabled. This suggests a latent trade-off between identity-preserving features and temporal shot precision within the VLM-encoder and DiT interface, a challenge we defer to future architectural optimization.

Secondly, we analyze relative consistency via Intra-Shot SC. While LTX-2-AV exhibits a high score of 0.87, this is largely an artifact of its failure to incorporate reference features, resulting in trivially static content with minimal frame variation. In contrast, our pipeline maintains a competitive Intra-Shot SC (0.6) while successfully grounding the target identity. This balance between dynamic visual content and identity fidelity further demonstrates the effectiveness of MTSS’s cross-shot reference grounding mechanism.

Thirdly, the implementation of identity customization does not degrade other performance dimensions. Metrics such as Sem. Following, A-V Sync, TA, and VQ remain consistently high across all MTSS-based variants. This stability underscores the successful decoupling of the factorized streams, proving that identity-specific information can be injected without interfering with the visual narrative or auditory synchronization.

\paragraph{Audio-Visual Synchronization.}
The audio-visual performance reveals a significant shift from low-information noise to semantically grounded content, driven by the explicit instructions in the MTSS Event Stream.

Firstly, the MTSS paradigm drastically improves speech accuracy and audio quality. As shown in Table~\ref{tab:auto-metrics}\ref{tab:human-eval}, LTX-2-AV-MTSS reduces the multi-shot WER from 0.84 to 0.13, a considerable imporvement achieved solely through representational replacement. This confirms that the factorized Event Stream provides the model with clear ``what to say'' instructions, effectively guiding the audio generator to produce semantically correct dialogue. Qualitative RMS analysis (Figs. \ref{fig:multi_shot_1}, \ref{fig:multi_shot_4}) substantiates this, showing that while the baseline produces flat, ambient-like envelopes, MTSS-based methods exhibit the rhythmic fluctuations characteristic of natural speech and action-driven events (e.g., periodic footstep impacts).

Secondly, we identify a synchronization tradeoff where the automated A-V Sync metric can be deceptive. While the baseline LTX-2-AV appears to lead with a Sync score of 6.86, its dismal human A-V score of 1.18 reveals that this advantage is an artifact of information sparsity; it is trivially easy to synchronize flat, ambient noise. In contrast, providing MTSS content initially increases the automated Sync score (to 13.86), as the model attempts to generate complex dialogue. However, our full pipeline successfully closes this timing gap, improving A-V Sync to 9.72 while maintaining a superior human A-V score of 2.26. 

Finally, we observe that foundational metrics—including Intra-Shot SC, Sem. Following, TA, and VQ—remain stable or exhibit further improvement across our MTSS-based variants. This empirical evidence validates the inherent logic of the MTSS paradigm: the shot and event streams are effectively factorized for independent semantic modeling, yet seamlessly integrated into a unified, coherent whole through reference and temporal grounding---which often enhances precise audio-visual synchronization.

% ============================================================
% Overall Summary
% ============================================================

\paragraph{Summary.}
Across all three axes and both evaluation paradigms, consistent conclusions emerge. The MTSS representational paradigm is the dominant driver of performance gains: with an identical architecture and training setup, simply replacing monolithic prompts with MTSS yields substantial improvements in every evaluated dimension---multi-shot controllability, cross-shot identity consistency, and audio-visual synchronization. Each factorized stream plays a distinct and critical role: the Shot Stream provides the temporal-segmentation prior that makes multi-shot control possible, the Reference Stream establishes persistent identity anchors that dramatically reduce character drift, and the Event Stream supplies explicit audio-event descriptions that transform speech and sound-effect generation from near-random to semantically accurate. Lightweight model adaptations---shot-aware attention, reference image injection---further amplify these gains by exploiting the learnable control interface that MTSS naturally provides. Importantly, these capability modules compose without mutual interference: our approach achieves the best or near-best performance across nearly all metrics in both automated and human evaluations.

\subsubsection{Training Details}
We adopted a phased training strategy to progressively master the structured components of the MTSS. The process began with staged optimization for individual tasks. \textit{ID Customization.} Performed for 3 epochs on a 400K identity-centric dataset using 128 GPUs to anchor persistent character and environmental traits.
\textit{Multi-shot Control.} Fine-tuned for 1.5 epochs across 250K multi-shot sequences on 32 GPUs to refine the shot-aware cross-attention mechanism.
\textit{Audio-Visual Synergy.} Executed for 3 epochs on 870K cinematic pairs using a 400-GPU cluster, specifically targeting the sub-frame alignment of the Gemma Connectors. Following the staged optimization, we conducted a final joint fine-tuning phase for 15K steps on 64 GPUs. This stage utilized an interleaved dataset comprising 60K high-fidelity cinematic alignment pairs and 250K multi-shot sequences. The objective was to harmonize identity stability, narrative coherence, and temporal synchronization within a single generative pass, ensuring these diverse constraints are satisfied simultaneously without mutual interference.

%% file: tables/caption_quality.tex
\begin{table}[t]
\centering
\small
\caption{Performance comparison on audiovisual captioning benchmarks. Evaluation is conducted using Gemini-2.5-Pro as the judge model. The thick line separates proprietary models (top) from open-source models and baselines (bottom).}
\resizebox{\textwidth}{!}{
\begin{tabular}{lcccccccc}
\toprule
\multirow{2}{*}{\textbf{Model}} & \multicolumn{4}{c}{\textbf{video-SALMONN-2 testset}} & \multicolumn{4}{c}{\textbf{UGC-VideoCap}} \\
\cmidrule(lr){2-5} \cmidrule(lr){6-9}
 & Miss $\downarrow$ & Incorrect $\downarrow$ & Hallucination $\downarrow   $ & Total Error $\downarrow$ & Visual Avg $\uparrow$ & Audio Avg $\uparrow$ & Details Avg $\uparrow$ & Overall $\uparrow$ \\
\midrule
\multicolumn{9}{l}{\textit{Proprietary Models}} \\
\midrule
Gemini2.5-Pro & 0.1902 & 0.0848 & 0.1209 & 0.3959 & 95.21 & 93.12 & 93.56 & 93.97 \\
Gemini2.5-Pro-MTSS & \textbf{0.1285} & \textbf{0.0644} & \textbf{0.0582} & \textbf{0.2511} & \textbf{97.33} & \textbf{93.54} & \textbf{95.65} & \textbf{95.51} \\
Qwen3.5-Omni-Flash & 0.2818 & 0.1175 & 0.1224 & 0.5217 & \textbf{87.85} & 65.92 & 74.46 & 76.07 \\
Qwen3.5-Omni-Flash-MTSS & \textbf{0.2104} & \textbf{0.0862} & \textbf{0.0688} & \textbf{0.3655} & 85.96 & \textbf{82.84} & \textbf{81.04} & \textbf{83.28} \\
\midrule
\multicolumn{9}{l}{\textit{Open-Source Models}} \\
\midrule
AVoCaDO & 0.2019 & 0.1211 & 0.1307 & 0.4537 & 84.72 & 84.12 & 80.22 & 83.02  \\
ASID-Captioner-7B & 0.2410 & 0.1324 & 0.1342 & 0.5076 & 83.48 & 77.38 & 76.52 & 79.13  \\
Qwen3-Omni & 0.3734 & 0.1015 & \textbf{0.1104} & 0.5853 & 72.81 & 52.81 & 62.76 & 62.80 \\
Qwen3-Omni-MTSS & 0.2831 & \textbf{0.0980} & 0.1355 & 0.5156 & 74.20 & 68.79 & 71.64 & 71.54 \\
Qwen3-Omni-MTSS-FT & \textbf{0.1548} & 0.1246 & 0.1119 & \textbf{0.3913} & \textbf{86.89} & \textbf{85.77} & \textbf{82.67} & \textbf{85.11} \\
\bottomrule
\end{tabular}
}
\label{tab:cap_quality}
\end{table}

\begin{table*}[t!]
\centering
\small
\caption{Quantitative results on audiovisual reasoning benchmarks. Gemini-2.5-Pro is employed as the judge model.}
% \resizebox{\columnwidth}{!}{
\begin{tabular}{lcc}
\toprule
\textbf{Model} & \textbf{Daily-Omni} $\uparrow$ & \textbf{WorldSense} $\uparrow$ \\
\midrule
\multicolumn{3}{l}{\textit{Proprietary Models}} \\
\midrule
Gemini2.5-Pro & 0.6825 & 0.4332 \\
Gemini2.5-Pro-MTSS & \textbf{0.7568} & \textbf{0.4981} \\
Qwen3.5-Omni-Flash & 0.3771 & 0.2476 \\
Qwen3.5-Omni-Flash-MTSS & \textbf{0.6145} & \textbf{0.3304} \\
\midrule
\multicolumn{3}{l}{\textit{Open-Source Models}} \\
\midrule
AVoCaDO & 0.5572 & 0.3531 \\
ASID-Captioner-7B & 0.5856 & 0.3203 \\
Qwen3-Omni & 0.1806 & 0.1569 \\
Qwen3-Omni-MTSS & 0.4117 & 0.3106 \\
Qwen3-Omni-MTSS-FT & \textbf{0.5945} & \textbf{0.3875} \\
\bottomrule
\end{tabular}
% }
\label{tab:cap2qa}
\end{table*}

%% file: 5_conclusion.tex
\section{Conclusion}

In this paper, we introduce Multi-Stream Scene Script (MTSS), a novel representational paradigm that replaces monolithic narrative paragraphs with deeply structured, factorized, and explicitly grounded scene descriptions. Through Stream Factorization and Relational Grounding, MTSS disentangles complex audio-visual dynamics into specialized streams and reconnects them via persistent identity and temporal links, fundamentally resolving the fidelity and scalability bottlenecks inherent in monolithic captions. Extensive evaluations demonstrate that this structured approach yields both more faithful video understanding and superior logical reasoning. Crucially, MTSS proves to be a significantly more learnable interface, narrowing the performance gap between smaller and larger MLLMs by relieving the burden of monolithic text processing. Beyond understanding, MTSS serves as a highly scalable control interface for downstream generation, substantially improving cross-shot identity consistency and temporal synchronization in multi-shot video synthesis. Ultimately, MTSS establishes a robust, data-centric foundation that advances video captioning into a genuine semantic interface, bridging raw multimodal inputs and complex narrative logic.

\paragraph{Limitations and Future Work.} While our study demonstrates the superiority of the MTSS paradigm in descriptive fidelity and reasoning, several challenges remain for future exploration. \textit{On the understanding side}, although our fine-tuning experiments show that MTSS-based learning significantly boosts performance, current open-source MLLMs still exhibit limitations in precise temporal localization, robust ASR performance, and accurate audiovisual entity-event association. Generating an accurate, deeply structured script requires exceptionally high cross-modal comprehension of the foundation model. Empowering more compact open-source architectures to attain the sophisticated scripting capabilities of proprietary models, such as Gemini, while effectively mitigating hallucinations, remains a formidable and unresolved research frontier.
\textit{On the generation side}, our exploratory experiments confirm that the structured format of MTSS is naturally suited for modern Diffusion Transformers (DiTs) in tasks like multi-shot control, identity customization, and joint audio-visual synthesis. However, we have only begun to scratch the surface of the synergy between MTSS instructions, VLM encoders, and DiT latent spaces. Further optimizing the architectural alignment between structured semantic priors and the generative process—to fully unlock the performance potential of MTSS—will require massive data scaling and more sophisticated cross-modal alignment strategies.